\documentclass[conference]{IEEEtran}
\usepackage{blindtext}
\usepackage{graphicx}
\usepackage{lipsum}
\usepackage{placeins}
\usepackage{bbm}
\usepackage{amsmath}

\usepackage{siunitx}
\usepackage{cite}

\ifCLASSINFOpdf
\else
\fi
\usepackage{algorithmic}

\usepackage{array}

\begin{document}
%
\title{Exploiting the Short-term to Long-term Plasticity \\ Transition in Memristive Nanodevice \\ Learning Architectures }



%
%

\author{\IEEEauthorblockN{
Christopher~H.~Bennett\IEEEauthorrefmark{1}, Selina~La~Barbera\IEEEauthorrefmark{2}, Adrien~F.~Vincent\IEEEauthorrefmark{1}, Jacques-Olivier~Klein\IEEEauthorrefmark{1}, \\
Fabien~Alibart\IEEEauthorrefmark{2}, and Damien~Querlioz\IEEEauthorrefmark{1}}

\IEEEauthorblockA{\IEEEauthorrefmark{1}Institut d'\'{E}lectronique Fondamentale, Univ. Paris-Sud, CNRS, 
91405 Orsay, France \\ 
Email: christopher.bennett@u-psud.fr}
\IEEEauthorblockA{\IEEEauthorrefmark{2}Institut d'\'{E}lectronique, Micro\'{e}lectronique et Nanotechnologies, UMR CNRS 8520, Villeneuve d'Ascq, France.}
}

%
%

%

\maketitle

\begin{abstract}
Memristive nanodevices offer new frontiers for computing systems that unite arithmetic and memory operations on-chip. Here, we explore the integration of electrochemical metallization cell (ECM) nanodevices with tunable filamentary switching in nanoscale learning systems. Such devices offer a natural transition between short-term plasticity (STP) and long-term plasticity (LTP). In this work, we show that this property can be exploited to efficiently solve noisy classification tasks. A single crossbar learning scheme is first introduced and evaluated. Perfect classification is possible only for simple input patterns, within critical timing parameters, and when device variability is weak. To overcome these limitations, a dual-crossbar learning system partly inspired by the extreme learning machine (ELM) approach is then introduced. This approach  outperforms a conventional ELM-inspired system when the first layer is imprinted before training and testing, and especially so when variability in device timing evolution is considered: variability is therefore transformed from an issue to a feature. In attempting to classify the MNIST database under the same conditions, conventional ELM obtains 84\% classification, the imprinted, uniform device system obtains 88\% classification, and the imprinted, variable device system reaches 92\% classification. We discuss benefits and drawbacks of both systems in terms of energy, complexity, area imprint, and speed. All these results highlight that tuning and exploiting intrinsic device timing parameters may be of central interest to future bio-inspired approximate computing systems. 
\end{abstract}


\section{Introduction}
Memristive nanodevices are a novel form of electronic memory whose properties are reminiscent of biological synapses, and which can exhibit plasticity features. In recent years, a variety of approaches have  been considered to integrate these devices into learning circuits; typically, a crossbar that stores analog weights is paired with a set of neurons built with CMOS technology. Collectively, these components manifest hardware systems that implement rules to perform learning tasks such as classification. These rules are often  bio-inspired, such as spike-timing dependent plasticity (STDP\cite{jo2010nanoscale,saighi_plasticity_2015,querlioz_2015}; elementary machine learning approaches such as the perceptron or gradient descent have also been implemented successfully \cite{backprop,onchiplearning}. Small demonstrator circuits show the physical feasibility of these rules \cite{alibart2013pattern, prezioso2015training}.

Intrinsic short term dynamics of memristive devices are not exploited in the aforementioned learning algorithms; rather, it is the sequence
of operations and implied long term plasticity that is crucial
to successful learning. However, just as synapses in the brain, some memristive nanodevices present rich plasticity behaviors over shorter term time scales \cite{chang2011short,wang2012synaptic,saighi_plasticity_2015,la2015filamentary}. In the brain, short-term plasticity (STP) refers to synaptic state change (potentiation) connecting neurons on the scale of seconds to minutes, while long-term plasticity (LTP) potentiates synapses for hours, days or even for a lifetime.  Although the transition from short to long-term plasticity is already considered in neuroscience models, such as meta-plastic learning systems \cite{abraham2008metaplasticity}, this transition remains an under-explored topic in the field of memristive learning due to an interest in the non-volatility of devices. While \cite{chang2011short,wang2012synaptic} explored memristive STP/LTP transitions and notably confirmed that repeated rehearsal of patterns evinces a strong analogy to the biological transition, they did not consider learning architectures based on these mechanisms. In \cite{burger2014volatile}, volatile tungsten-based memristive devices that relax relatively quickly were considered for integration into a learning system, but only the STP regime exploited for classification.

Using the transition from short-term to long-term plasticity as the core component of a nano-electronic learning system is therefore a novel approach. Here, we explore the merit of that approach  by constructing two learning architectures and testing them on two classification tasks. We focus on a highly promising device, electrochemical metallization (ECM) cells where rich STP and LTP dynamics have been evidenced recently \cite{la2015filamentary}. First, we introduce the time dynamics of ECM nanodevices and an experimentally validated model of their behavior. Second, a single crossbar approach based on this concept is proposed and simulated. To overcome its limitations, we subsequently introduce a system partly inspired by Extreme  Learning Machines (ELM) that exhibits exciting performance and high resilience to device variability.  This allows us to finally discuss merits and drawbacks of our approach.


\section{Nanodevice Plasticity Model}
\label{plasticity}
\begin{figure}[h]
\centering
\includegraphics[width=\columnwidth]{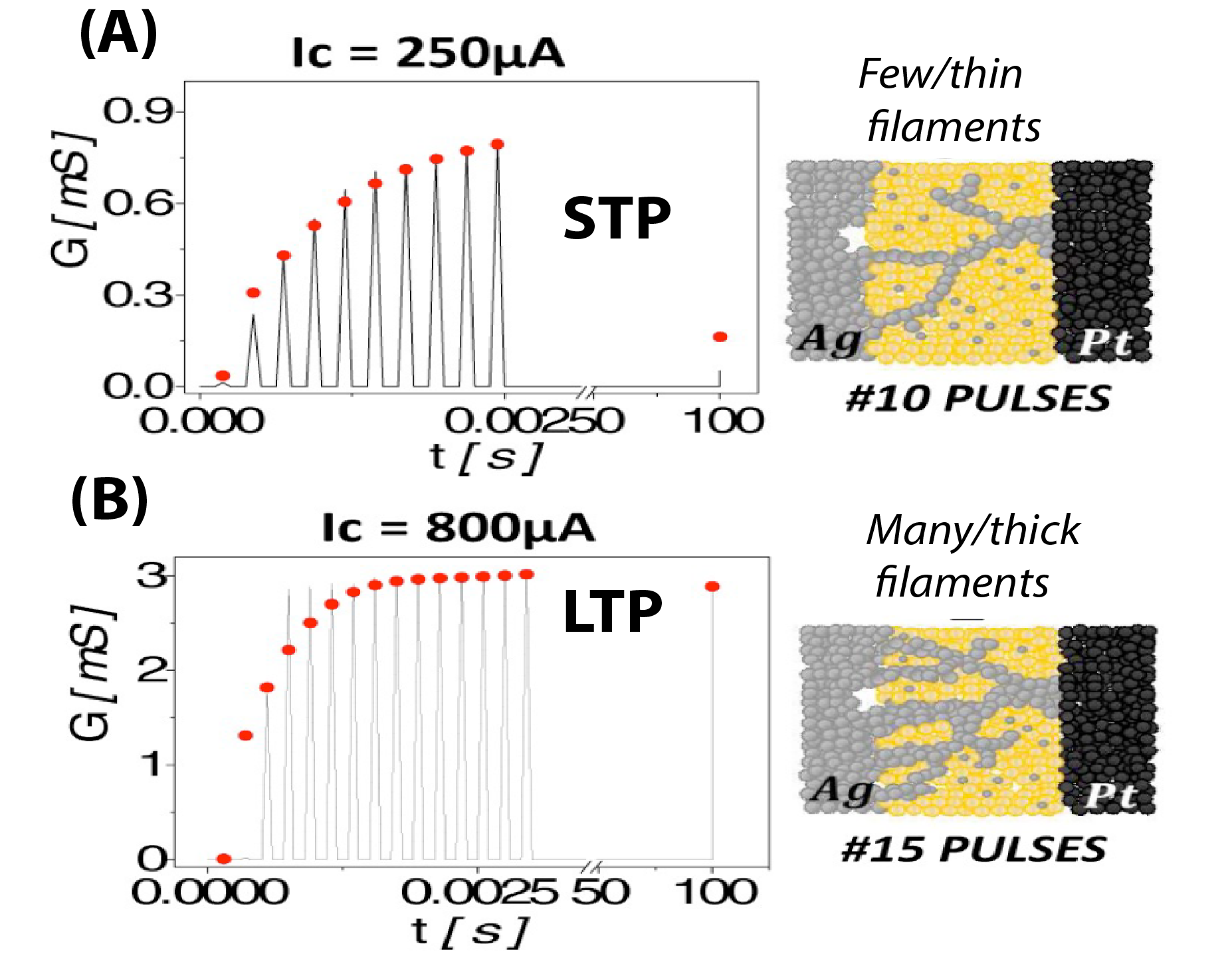}
\caption{(A) depicts a pre-synaptic spike train that keeps a device in the STP regime; a corresponding, weaker filament that can easily relax is pictured. (B) depicts a more powerful spike train that successfully moves the pictured device from the STP to LTP regime. The black solid lines are the measurement results and the red dots are the model predictions.}
\label{filament}
\end{figure}

The devices considered are electrochemical metallization (ECM) cells with a 60 nm switching layer, where dendritic filaments form in between a reactive top electrode (anode) of silver, and an inert bottom electrode (cathode) of platinum \cite{la2015filamentary}. The application of a positive bias above a threshold $V_\text{th}$ causes oxidation and drift of silver ions ($Ag^{+}$) across the $Ag_{2}S$ switching layer from the cathode towards the anode. This increases conductivity and physically corresponds to the formation and strengthening of filaments. Conversely, a negative bias induces reduction at the $Ag$ electrode, weakening filaments and decreasing conductivity. The device shows a natural relaxation towards lower conductivity as $Ag^{+}$ ions continue to diffuse and reverse oxidation-reduction occurs. Critically, this natural relaxation may be fast or slow, depending on the quantity and quality of the filaments. As reported in \cite{la2015filamentary}, varying filamentary morphology and a possible trade-off between filament density and diameter create complex synaptic behavior. In particular, the transition from a relatively small relaxation time ($\tau$)- the STP regime-  to a larger $\tau$ corresponding to the LTP regime - was tunable both by the number and the characteristics of subsequent pre-synaptic excitatory pulses.
Fig.~\ref{filament}(A) depicts the STP case where a small number of pulses strengthen the filament so that the ECM cell's conductance increases to $0.9 mS$. However, this state is not stable: after a time $\tau = 100 s$, the conductance has relaxed to a low conductance state. Conversely, Fig.~\ref{filament}(B) depicts the LTP case. 
Differently timed spikes move the synapse to a high conductance of $3 mS$, which remains stable after $\tau = 100 s$. 

A detailed model of STP to LTP transition in ECM cells, reminiscent of a biological model of plasticity \cite{markram1998potential}, was validated experimentally in our previous work \cite{la2015filamentary}. 
We now revisit the basic equations of this model.
Synaptic potentiation increases in response to a train of pre-synaptic pulses; the facilitating time constant $\tau_\text{fac}$ constantly increases as spikes are applied and conductance increases, facilitating the STP to LTP transition. After each programming spike:
\begin{equation}
\tau_\text{fac} = a \cdot G(t)^{b},
\label{taufac}
\end{equation}
and after any given delay $ \Delta t $ from the last spike at time $t$, the conductance is
\begin{equation}
G_\text{relax} = G(t) \cdot \exp{\left(\frac{-\Delta t}{\tau_\text{fac}}\right)}
\end{equation}
Finally, at time $t + \Delta t$, the conductance value results from the sum of the exponential relaxation and of a programming spike if any is applied:
\begin{equation}\label{gevolve}
	G(t + \Delta t) = 
	\begin{cases} 
	G_\text{relax} + U(A-G_\text{relax})	& \text{, if spike} \\
	G_\text{relax} 						& \text{, if no spike}. \\
	\end{cases}
\end{equation}

$A$ corresponds to maximum synaptic efficiency ($G_\text{max}$). A typical value extracted from device measurement is $A = {4}{\milli\siemens}$. The typical synaptic efficiency is $U = 0.025$. The power law prefactor $a = \SI{2.42e-12}{\s \per \siemens \tothe{b}}$, and the power law exponent $b=4$ \cite{la2015filamentary}. 

\section{A Simple Learning Task and Algorithm}
\label{sec:simple}

We first introduce a simple architecture that highlights the promise and challenge associated with exploiting the STP to LTP transition in nanodevices.

\subsection{Crossbar Architecture and Algorithm}
The architecture  has three components: a software image database or sensor and circuitry to convert pixels into voltage spikes, an all-to-all crossbar that connects input and output neurons electrically at each ECM cell crosspoint, and accompanying CMOS circuitry. Learning occurs in three stages. 

\begin{figure}[h]
\centering
\includegraphics[width=\columnwidth,height = 12cm]{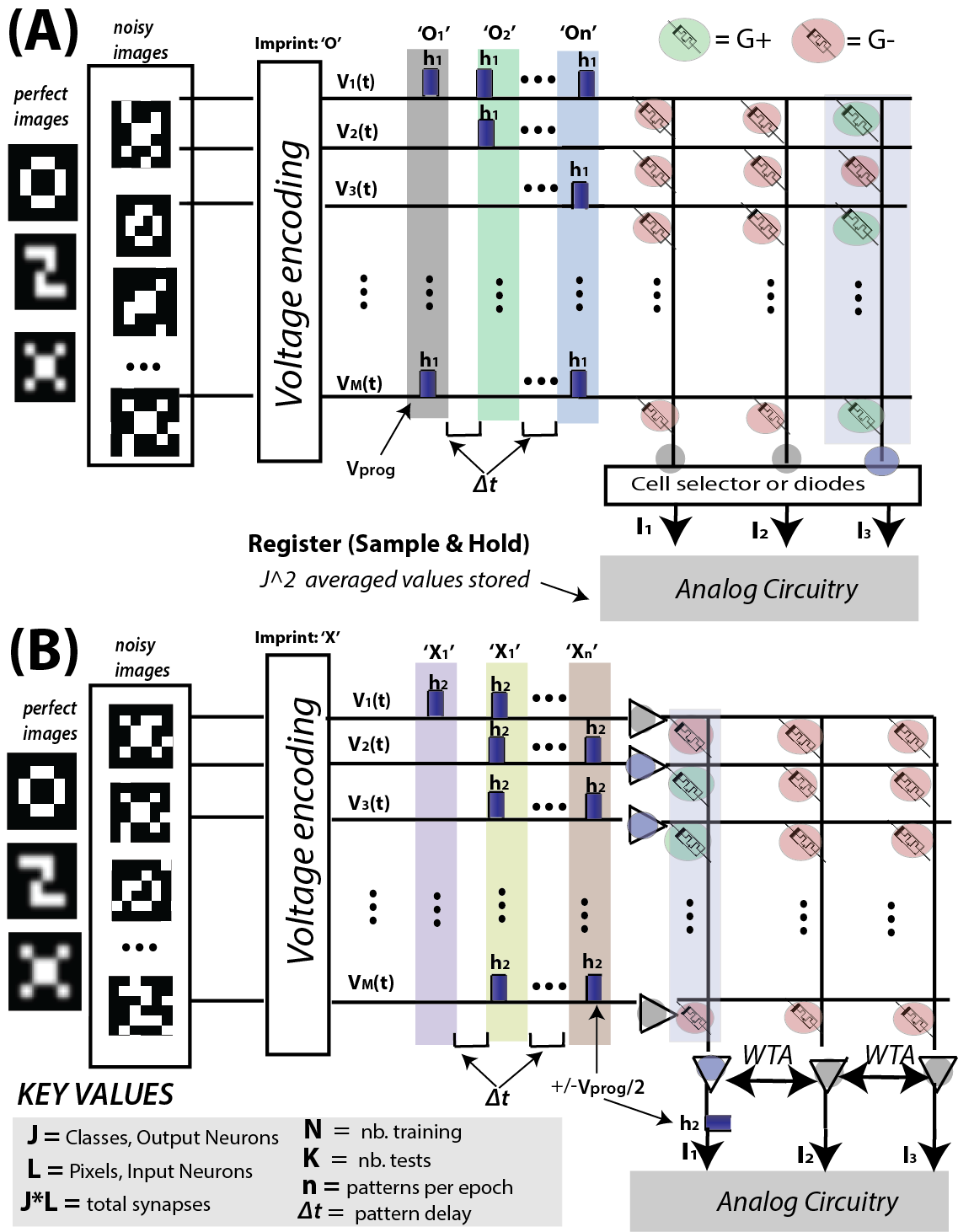}
\caption{The simple learning system, with characteristic images input to the system, selected and non-selected nanodevices, and input and output computing accessories required for learning. (A)  supervised scheme, and (B)  unsupervised scheme. Key variables are also noted. }
\label{schematicsimple}
\end{figure}

\begin{figure}[h]
\centering
\includegraphics[width=\columnwidth]{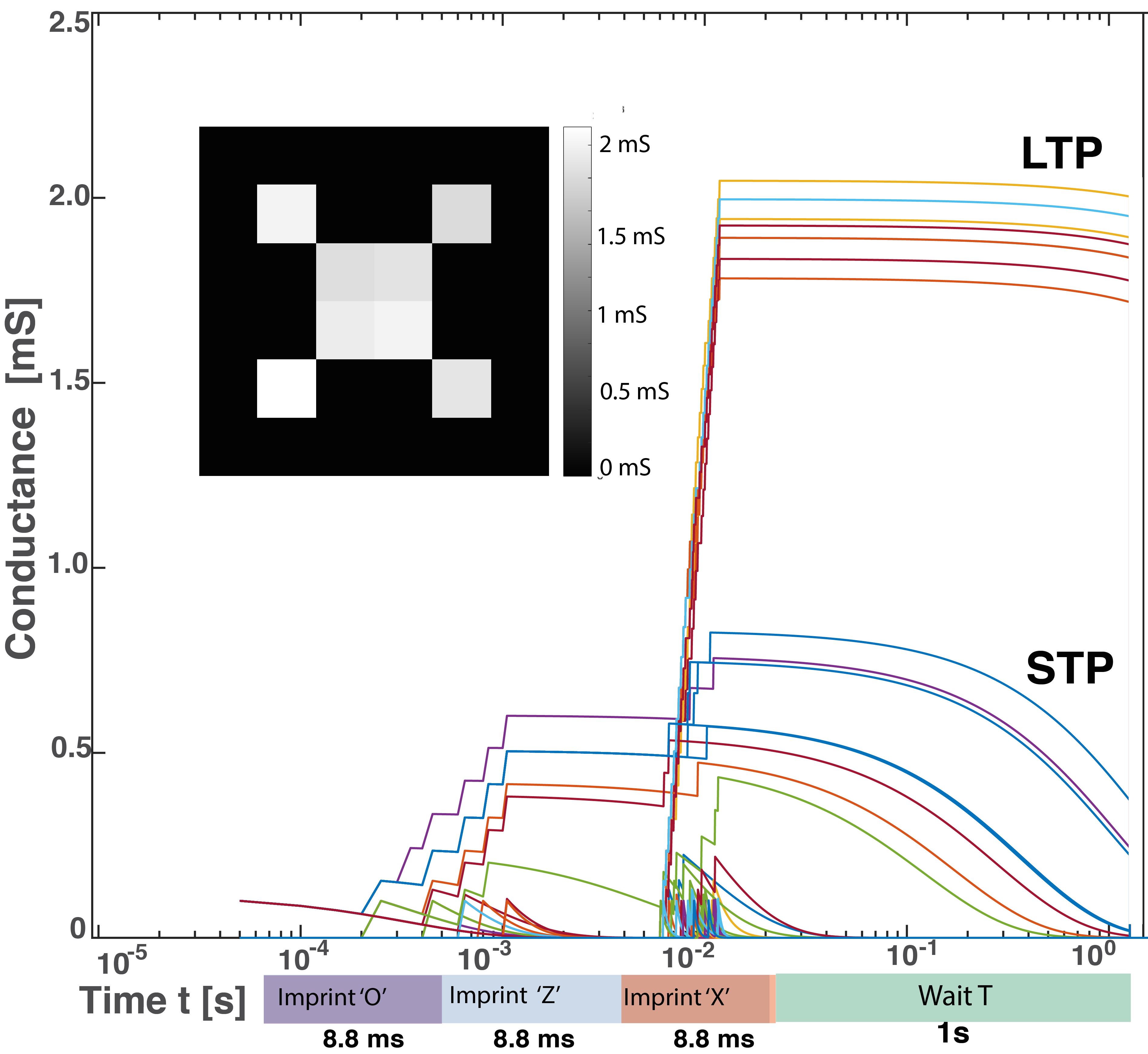}
\caption{Conductance as a function of time during the imprinting process- $n=30, \Delta t = 200  {\micro\second}$ and subsequent waiting period, for the 36 ECM cells connected to a single output neuron. Inset: conductance of the cells at  t = \SI{1.026} {\s}, presented as a reconstructed 2-D image. Note time is portrayed on a semi-logarithmic and not linear axis.}
\label{stpltp}
\end{figure}

In the first stage, ECM cells are ``imprinted''. This stage is sub-divided into several epochs corresponding to the total number of classes $J$. In this case, $J=3$: classes 'O', 'Z' and 'X'. Each image contains $L=36$ total and 8 active pixels. During each epoch, $n$ noisy examples of the given class are subsequently presented to the crossbar such that a spike represents an active (white) pixel. Each input neuron receives one pixel consistently, and the patterns are presented with a delay $ \Delta t $ as depicted in Fig. \ref{schematicsimple}, where (A) is in epoch 'O' and (B) in epoch 'X'. After imprinting, no voltage is applied on the crossbar for a wait period $T$. This allows ECM cells in the STP regime, assumed to be noise, to return to low conductance states, while it will not affect those in LTP. Fig. \ref{stpltp} presents the evolution of the conductance of the 36 ECM cells connected to one particular neuron, during imprinting ($t_\text{imprint} = 26.4  {\milli\second}$) and subsequent wait ($T = 1s$). The final conductance map for the output neuron- in this case it has learned 'X'- is visualized pixel-by-pixel in the inset. 

Imprinting only works if each output neuron corresponds to a different class. Output neurons may be mapped to classes either in a supervised manner-  employing a cell selector such as the FAST selector \cite{jo20143d}- or in an unsupervised manner, such as a winner-takes all (WTA) half-select scheme. Only nanodevices at the intersection of an active pixel/row (receiving pre-synaptic spike $V_\text{prog,h} = 0.42V$, $V_\text{prog,w} = 100 {\micro\second}$) and the selected column (green nanodevices in Fig.~\ref{schematicsimple}(A),(B)) increase conductance. In the unsupervised scheme depicted in Fig.~\ref{schematicsimple}(B), input spikes are set at $V_\text{prog}/2$; only with a complementary spike of the 'winning' leaky integrate and fire (LIF) output neuron ($-V_\text{prog}/2$) are synapses in the appropriate column imprinted. Here, all other LIF neurons besides the one who spiked first are inhibited due to a lateral diffusion scheme 
\cite{vincent2014spin}. Conductance evolution is not equivalent between the two cases due to half-select effects,  yet transition from STP to LTP is nevertheless possible in both systems. The supervised scheme generates all following simulation results.

Second, images are presented to the network in 'read' mode ($V_\text{read} = 0.1V$) - so as to not disturb conductances- and currents are read out at all output neurons (not just the corresponding one). As output currents are a dot product of device conductances and active pixels for a given image, many unique values are possible. These values are stored in a circuit (register) below where they are iteratively averaged. After $N$ examples, $J^2 = 9$ currents (signatures) are stored: $I_\text{reg}$. Testing is the final phase: $K$ unknown digits are presented at $V_\text{read}$ and output currents $I_\text{test}$ are compared to the register's values. The predicted class is the one which minimizes $E_\text{tot}$:
\begin{equation}
E_\text{tot} =  \sum_{i=1}^{J} {\left| I_{reg_i}- I_{test_i} \right| }
\end{equation}
If predicted class is the true class, '1' is placed in a ledger; else '0'. The final score is simply ledger sum divided by $K$. Computing an iterative average during training and storing it in a register may be achieved in either analog (operation amplifiers, sample and hold circuits each containing a capacitor), or digital (analog to digital converters and conventional digital memory, eg RAM) fashions, or a combination. Computing $E_\text{tot}$ during tests additionally requires an absolute value circuit. Each output neuron (class) must have access to equivalent circuitry. Agnostic of implementation, the circuit overhead is non-negligible and a drawback of this approach.

All of the following reported results, as well as that depicted in Fig.~\ref{stpltp}, were produced by a software program 
that simulates a crossbar of nanodevices, each following the mathematical model for conductance evolution introduced in Section \ref{plasticity}, and tracks evolution of synapses and currents over time in response to voltage encoded input spike trains.
This simulation software also models nanodevice specific issues such as device variability. For the simulation results immediately following, $N=K=100$, long wait $T=1s$. Noise is added by randomly flipping to their opposite state \SI{10}{\%} of all pixels in images used for imprinting, training, and testing. 

\subsection{Performance on the Simple Task}


\begin{figure*}[!h]
 \center
\includegraphics[width=\textwidth, height = 13 cm]{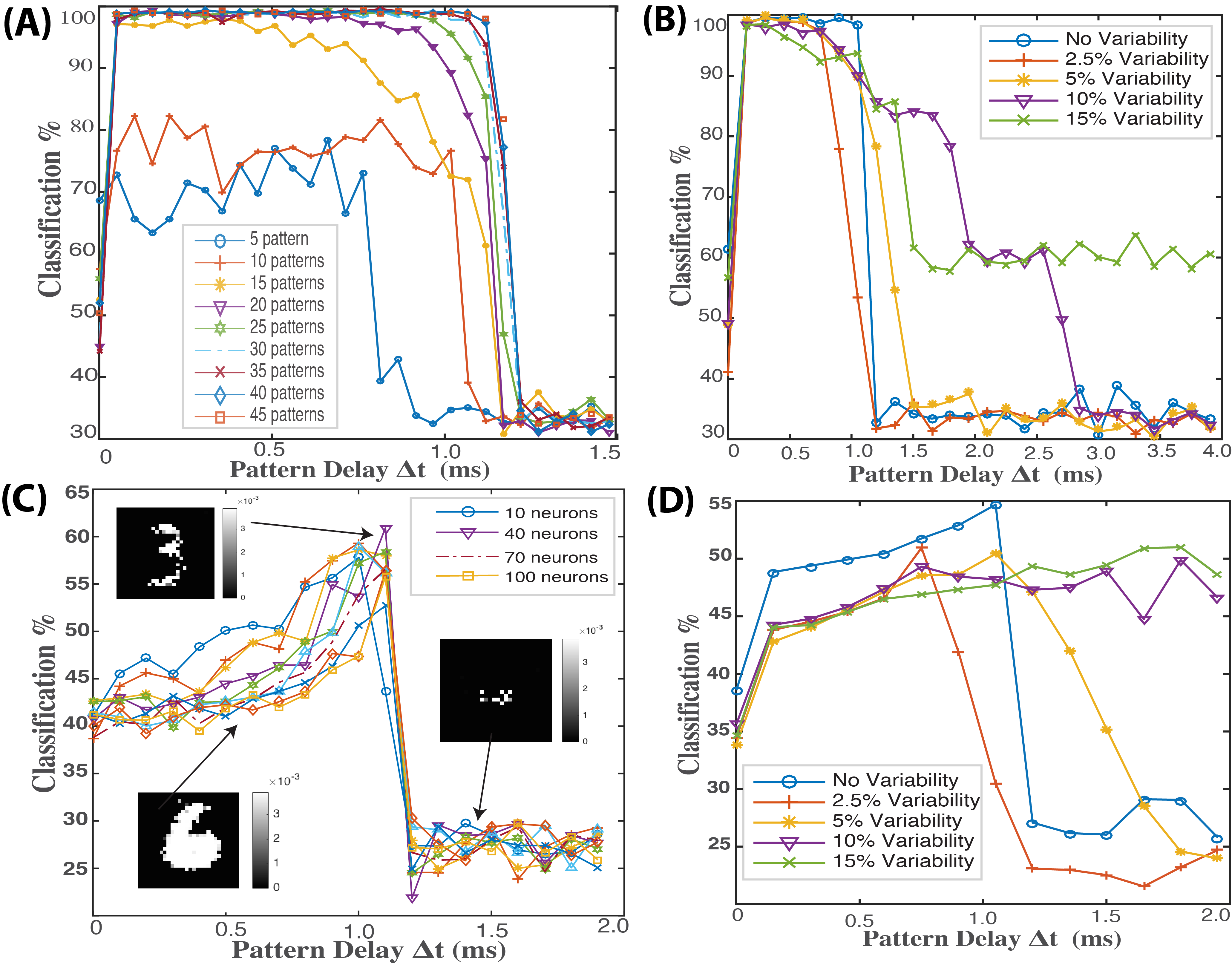}
\caption{Top Panels: Classification rate of the simple system on the simple images, as a function of time step $\Delta t$. (A) for different number of patterns presented per epoch (class) and (B) for different degree of device variability ($n=45$ patterns imprinted per epoch). Bottom panels: Classification rate on MNIST also as a function of $\Delta t$. (C) Varying $J$ output neurons when constantly $n=50$. The three insets depict reconstructed conductance maps for one of the ($J_i$) neurons imprinted at given $\Delta t$ parameter; white represents device in LTP (ON), and black those in STP (OFF) (D) Effect of degree of device variability ($J=10$ output neurons,$n=50$ in each case). Every image has 10\% noise.}
\label{simple}
\end{figure*}

Fig.~\ref{simple}(A) presents classification rates as a function of the chosen inter-pattern wait step $\Delta t$. Each series represents a different number of patterns $n$ presented per epoch of imprinting stage. The performance reaches a nearly perfect 98\% over a  broad timing range ($ 0.1 {\milli\second} < \Delta t < 1.2 {\milli\second} $) for $n>20$.  Below $n = 20$, it is not  possible to reach the plateau as there are insufficient presynaptic pulses to move nanodevices from the STP to LTP regime (also visible in Fig.~\ref{filament} and discussed in \cite{la2015filamentary}).

 Fig.~\ref{simple}(A) also shows sub-optimal classification before and after the optimal range. The former occurs when patterns imprint too fast to synchronize with the nanodevice's normal relaxation parameter $ \tau_{fac} $, hence the conductance map is over-saturated (too many devices enter LTP). Conversely, when $ \Delta t $ is too large, insufficient devices enter LTP to retain the digit image. In these cases currents no longer vary meaningfully neuron-by-neuron, and classification becomes difficult.

\subsubsection{Effect of Device Variability}
Nanodevices always suffer from some device variability \cite{querlioz_2015}. We consider the case where ECM cells each behave slightly differently to equivalent pre-synaptic spikes. Each now receives a different internal device timing variable: $U, A, a$. From  Eq.~\ref {taufac}, each device then possesses a slightly different $\tau_{fac}$ as $a$ changes; from Eq.\ref{gevolve}, each conductance evolves a bit differently due to varying synaptic efficiency ($U$) and $G_\text{max}$ ($A$). 
Random values $U, A, a$ are drawn from a normal distribution with mean ($\mu$) set as those listed in Section II, and coefficient of variation considered over degrees $ \sigma/\mu = \{ 0.025,\, 0.05,\, 0.1,\, 0.15\ \}$.  For each degree of variation, 20 simulations were performed at each $ \Delta t $ value and averaged. As depicted in Fig. \ref{simple}(B), increasing variability reduces the nearly perfect classification plateau. Unlike the uniform case, increasingly variable crossbars do not fall off a performance 'cliff', but experience a gentler landing at increasing dispersion parameters. In the $ \sigma/\mu = 0.15$ case, the same recognition 'floor' as in the other cases (~30\%) is not reached even at a very long inter-pattern wait ($4 ms$). 


\subsubsection{Effect of Increasing Training Samples}
Few examples are needed in order to learn the functions, because the register reaches a useful average very quickly. In the uniform case classification reaches 70\% with 5 samples, and approaches 100\% after 10. Considering variability, 15 samples are needed to reach peak (95\%) classification; however, the  highest dispersion case (15\%) takes 25 samples to reach its peak (90\% classification).

\subsubsection{Effect of Increasing Noise}
In the uniform case, classification remains nearly perfect until around 15\% of pixel flips and then deteriorates linearly after that until a minimum of 75\%. Low and medium variability cases perform well until 10\%, following a similar deterioration trajectory thereafter. However, the highest variability case (15\%) only does as well as the others in the 0-5 \% noise range; by 20\% noise-induced flips it already falls to 80\% correct. 
\begin{figure*}[!h]
 \center
\includegraphics[width=\textwidth, height = 10 cm]{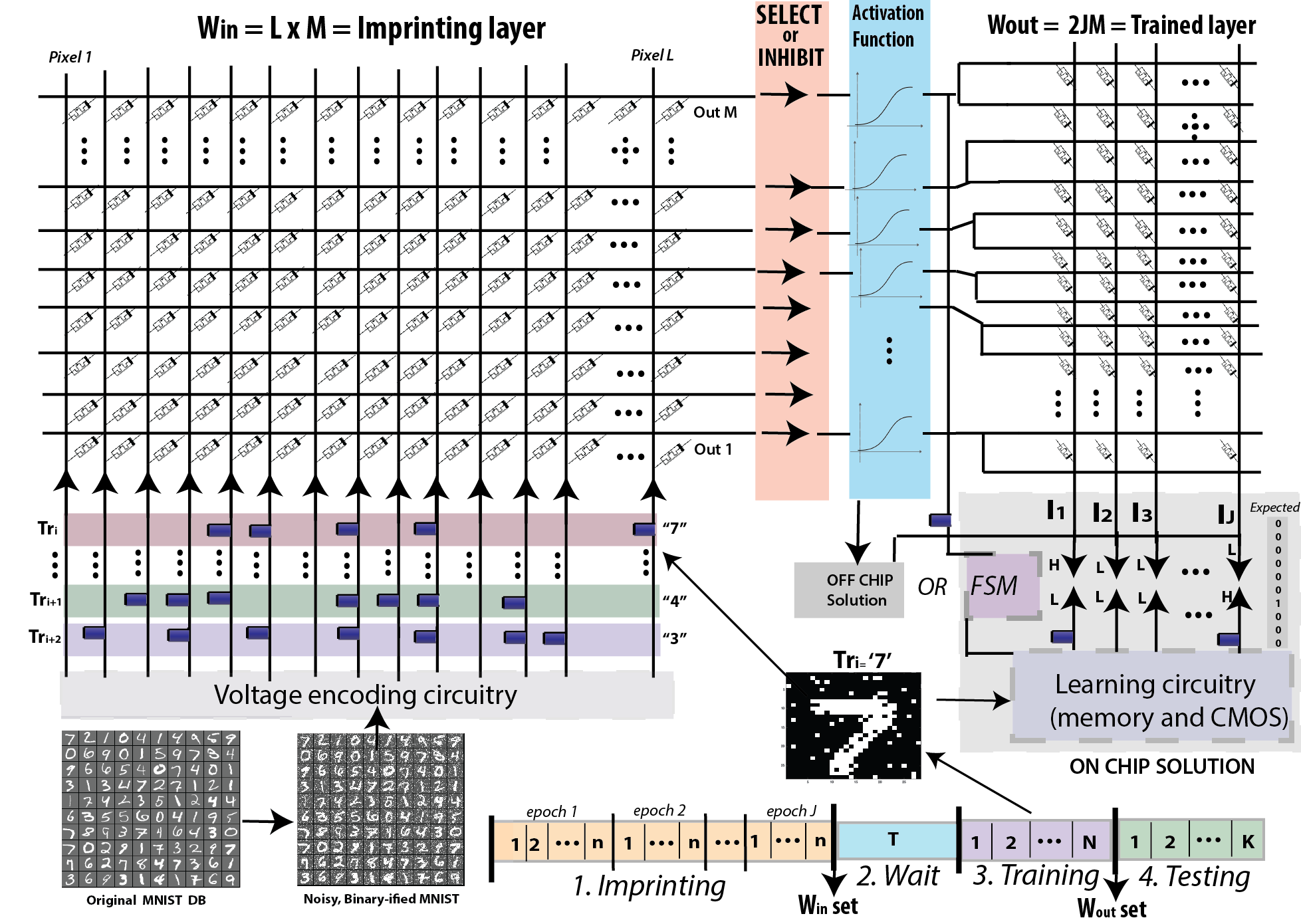}
\caption{Conceptual architecture diagram for the dual-crossbar system that can compute higher dimensional tasks by projecting currents from examples to a second regression layer. One representative moment of the system - presentation of an image $Tr_{i} = 7$ during the training period- is depicted. The timeline below depicts operation through major phases. All variables are the same as defined in Fig. 2.}
\label{schematicELM}
\end{figure*}
\subsection{Performance on the MNIST Task}
In attempting to classify the MNIST database of hand-written digits \cite{lecun1998gradient}, $J=10, M = 784$ so 7840 synapses (ECM devices) attempt to resolve the problem. In this case, the register must hold $J^2 =100$ values. While MNIST provides $N = 60k$ training, $K = 10k$ tests, only $N=K=1k$ were used. Overall, the system's performance on this task is not favorable. Fig.~\ref{simple}(C) shows that classification peaks at $ \Delta t =1.1 ms$ with 61\% correct. The insets in Fig.~\ref{simple}(C) highlight that at this peak, digits are relatively well constructed if sparse, while reconstructed pixel maps in the mostly evaporated (super-optimal $\Delta t$) and oversaturated (sub-optimal $\Delta t$) regions are unusable. However, even at the peak, classification is weak because of an intrinsic algorithmic weakness. Since all the register memorizes are currents and test images have a wide variety of active pixels (in contrast to the small images), it has a hard time distinguishing between different classes that produce similar current sums. Fig.~\ref{simple}(C) also suggests that increasing the number of output neurons beyond $J=10$ does not help with the present algorithm, as stored averages for redundant neurons will be similar. Fig.~\ref{simple}(D) shows that device variability has a  deleterious effect on learning ability. At low dispersion, degree peak classification drops to 50\% while preserving a similar trend, while large dispersion echoes the phenomena of resilience to large pattern delays observed in Fig.~\ref{simple}(B).

\section{An ELM-Inspired Approach}
\label{sec:ELM}
Inferior performance on the harder task, and complex readout scheme, inspired us to expand from a one-crossbar system. Rather than using currents from the imprinted layer to solve a classification problem, we considered the case where those currents are passed forward, after being transformed at the hidden layer via an activation function, to a second crossbar. We obtain a hardware instantiation of the principles of a single hidden layer feedforward network, which is reminiscent of the extreme learning machine (ELM) method \cite{huang2004extreme}.  

\subsection{System Description}

\begin{figure*}[!h]
 \center
  \includegraphics[width=\textwidth, height = 6cm]{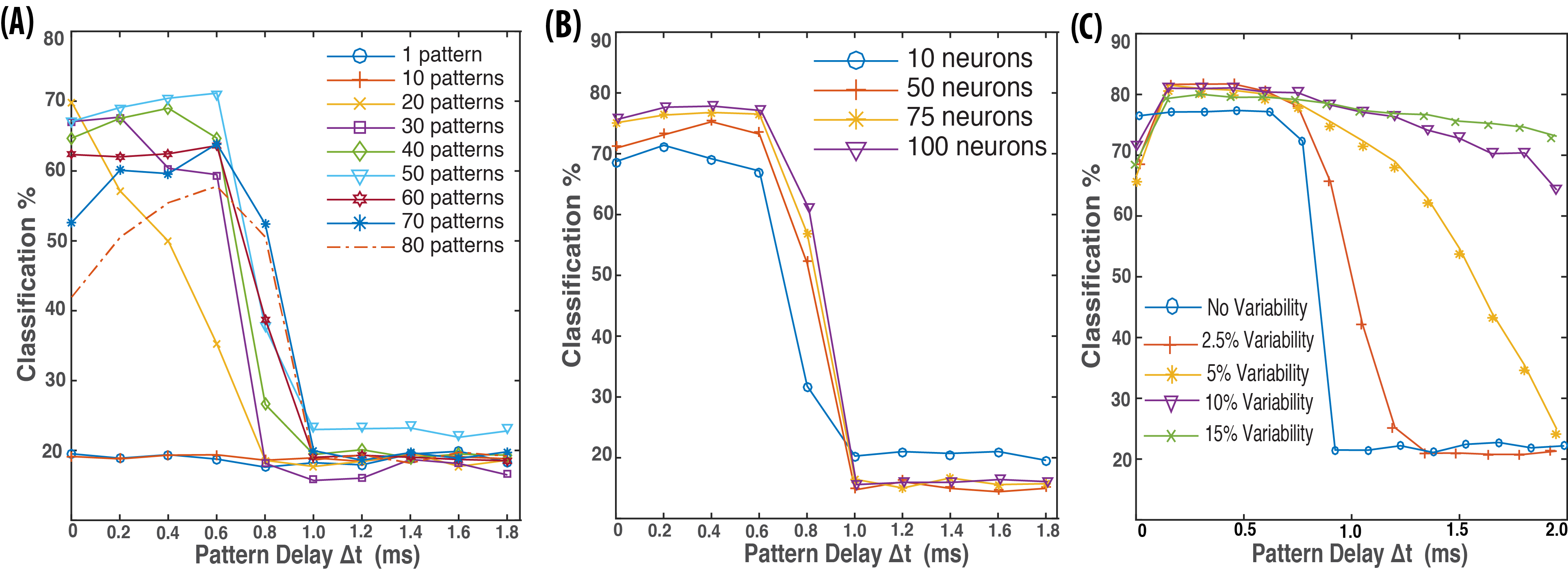}
  \caption{Classification rate as a function of delay between pattern presentation during the imprinting phase, for the ELM inspired system.
  (A) for different number of patterns presented per epoch (M=10)  (B)  for different number of hidden neurons M  (n=50 patterns) (C) for different nanodevice variability cases (M=100, n=50). In every case, $T=1s$, $N=60k$, $K= 10k$, 10\% noise applied to every image in imprinting, testing, training.}
  \label{ELMtiming}
\end{figure*}

Fig. \ref{schematicELM} reveals a conceptual hardware implementation of our ELM-inspired system built with two crossbars of memristive devices, along with a timeline of its operation. There are four phases: imprinting, waiting, training, and testing.

The original database is presented to the inputs of the first crossbar as binary, noisy, voltage vectors on demand. The first crossbar uses ECM cells as described in the previous section. It is imprinted with images taken from the training dataset,  using the same procedure as in section~\ref{sec:simple}. It is therefore the part of the system that uses the STP to LTP solution, and it yields a projection space \( W_{in} \) that will be used in training and testing. This is in contrast with conventional ELM where the weights of the first layer are random \cite{huang2006UniversalApprox}.
Unlike the system detailed in Section~\ref{sec:simple}, currents are not stored but fed to the the second stage of the system. Before they reach the second crossbar, they are passed through activation functions to increase dimensionality. In the ELM scheme, each activation function is slightly different. The \( tanh(I) \) function was chosen since it can be easily implemented in CMOS and engineered for variability (offset and/or gain factor). In our case, offsets were set randomly on each neuron and gain factor always held at 10. Imprinting happens epoch by epoch and the total number of epochs is identical to size of the hidden layer, $M$.


The second layer's weight matrix \( W_{out} \) acts as a regression layer, and is not imprinted but trained. It does not exploit the STP to LTP transition of the nanodevices. Each input of the second crossbar is connected to two rows of the second crossbar. This allows us to model positive as well as negative weights connecting input and output \cite{onchiplearning, onchipuniversal,bennett2015supervised,chabi2015ultrahigh}. A least squares solution may be obtained by computing weights at the end of training and  importing them (batch mode), or may be computed iteratively as subsequent training examples are given (online mode).
Batch solutions include a pseudo-inverse operation, which might be complex to compute in hardware,  and closed-form ridge regression, which may be easier to implement. A pseudo-inverse computed online can achieve promising classification \cite{van2015online}, while iterative on-chip training schemes especially for memristive nanodevices exist and can be naturally implemented in a memristive crossbar. In this scheme, each column or class of the second (regressed) layer implements its own approximation of the Widrow-Hoff algorithm and they can be programmed parallely  \cite{onchiplearning, onchipuniversal,bennett2015supervised,chabi2015ultrahigh}. Both options are visualized in Fig. \ref{schematicELM}. Batch and online options  will be compared extensively in a follow-up paper. For consistency, the results presented hereafter always used batch learning (closed-form ridge regression) to obtain \( W_{out} \) given actual matrix $A$ of training examples (composed of projected current vectors from \( W_{in} \)), and expected matrix $Y$ (composed of binary vectors of presented classes):
\begin{equation}
W_{out} = Y A^\top inv(A A^\top) 
\label{inv}
\end{equation}
In previous proposals for implementing ELM with memristive devices, the weights of the first layer are random, using the intrinsic nature of device variability- in particular, variance in the OFF state of the memristor \cite{ELMMemristor}. Here, we instead harness an imprinting made possible by the STP to LTP transition. We include a direct comparison of this new approach with the random conductance values approach in the next subsection. 

\subsubsection{Effect of Device Timing and Variability}
 Fig.~\ref{ELMtiming}(A) again shows that a sufficient number of patterns presented per epoch (imprinted neuron) $n > 20$ is a constraint for successful imprinting. Conversely, over-saturation is also possible when $n > 50$ patterns are applied per epoch at the faster (smaller) time steps. In all uniform cases, performance drops when imprinting is too slow ($ \Delta t > 1 {\milli\second}$). Fig.~\ref{ELMtiming}(B) shows that unlike the simple system, increasing output (hidden layer) neurons increases performance.  This is due to the different random activations provided at each neuron.  Whereas Fig.~\ref{simple} showed a reduction in performance at increasing dispersion, Fig.~\ref{ELMtiming}(C) reveals the contrary case: maximum classification slightly increases. For instance, when $M=100$ and at optimal wait, max 78\% is reached in the uniform case compared to 82\% for variable. While the uniform case shows a classification 'cliff' after $ 0.8 {\milli\second}$, the 10\%, 15\% variable cases again show a broad tolerance to slower imprinting. One explanation is that, with high conductance evolution variance, some synapses are always excited enough to move from STP to LTP. This  result is attractive, since nanodevice variability is transformed from a liability into a productive asset of the computing system.
 \subsection{Performances}
\label{sec:hiddensize}
 \begin{figure}[h]
\centering
\includegraphics[width=\columnwidth]{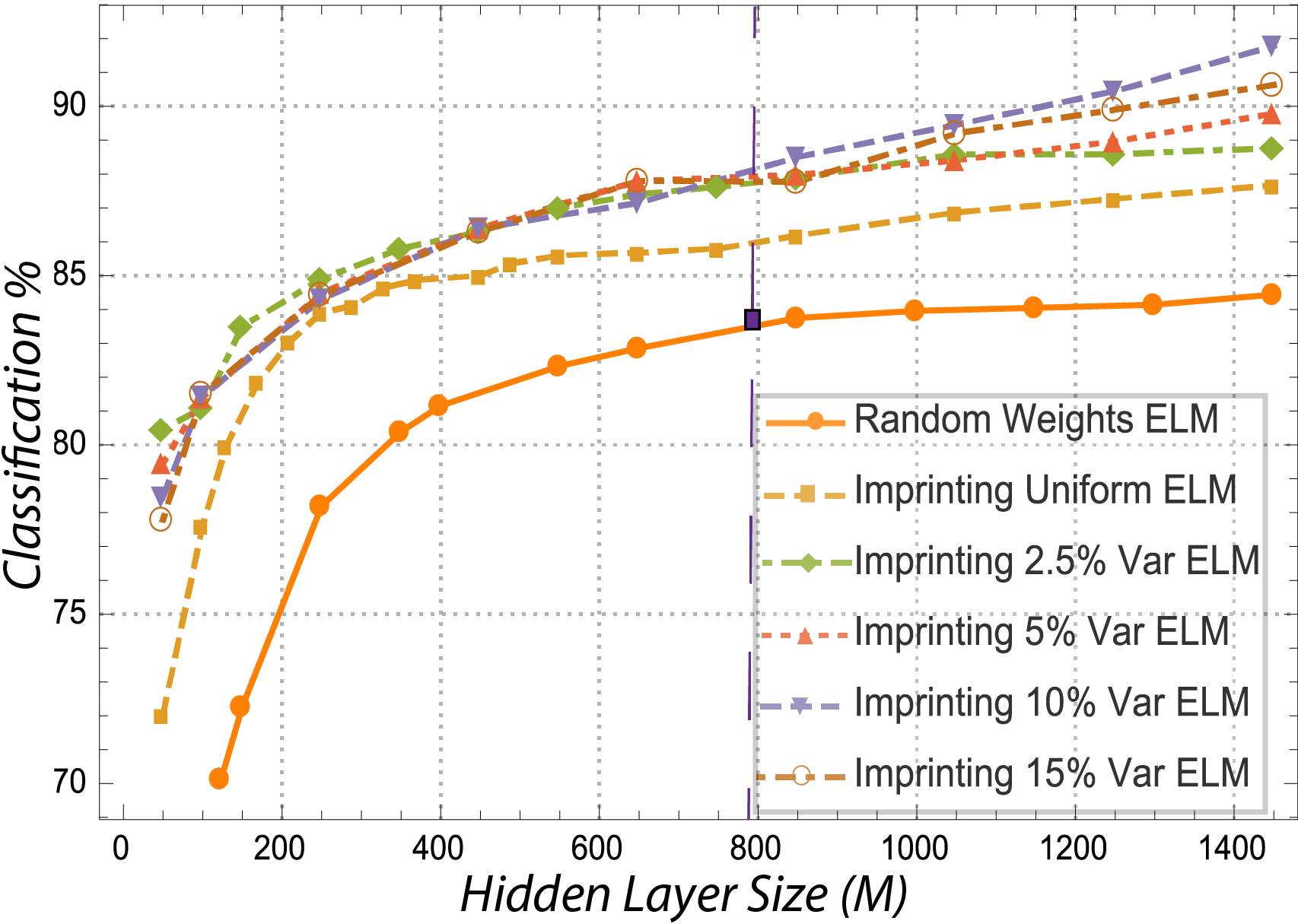}
\caption{Classification rate as a function of hidden layer size M in different conditions: random weights on first layer (Random Weights ELM); imprinting on first layer, with no variability (Imprinting Uniform ELM); or variability (Imprinting Var ELM) on nanodevices. In every case: $n=50$ patterns are given per epoch, $T=1s$,  $ \Delta t = 200 \mu s$, 10\% noise is present in every imprinting, train, and test image, $N=60k$,$K=10k$, each hidden neuron slightly varies its activation function with a gain factor constantly 10. The single purple point/ dashed line at  $L=M$ represents the direct regression solution obtained when all training images are presented directly to the second layer without any first layer (projections).}
\label{HiddenLayer}
\end{figure}

\subsubsection{Effect of Hidden Layer}

Fig.~\ref{HiddenLayer} shows that regardless of device uniformity or variability, imprinting \( W_{in} \) is demonstrably meaningful: imprinted systems substantially out-perform the ELM control cases at every value of $M$. 
This result can be compared with analogous priming of the first layer of ELM systems in software artificial neural networks. Such priming has already been reported to improve performance over the standard case \cite{mcdonnell2015fast,tapson2015explicit}. Here, we show a similar result subject to unique device timing constraints. Fig.~\ref{HiddenLayer} also shows that imprinted systems with synaptic evolution variability consistently outperform the uniform case (where every synapse behaves identically). Nanodevice variability then allows for a greater variability between the hidden neurons, enhancing the dimensionality of the data provided to the second layer beyond just the varying activation functions. At $M=1450$, peak classification of 91.8\% is obtained for variable imprinted systems, the uniform imprinted systems obtain $87.8\%$, and random weight ELM reaches 84.4\%. At $M=L=784$ (dashed vertical line in Fig.~\ref{HiddenLayer}), random weight ELM performs similarly to a regression obtained by presenting all training samples directly to the second layer (83.5\%). While the direct regression can only be made at $M=L$, to reach higher performances than standard regression, ELM requires substantially higher M values.  Conversely, the rich underlying dynamics of nanodevices allows designers to do more with less in the imprinting cases.


\subsubsection{Effect of Training Set}

\begin{figure}[h]
\centering
\includegraphics[width=\columnwidth]{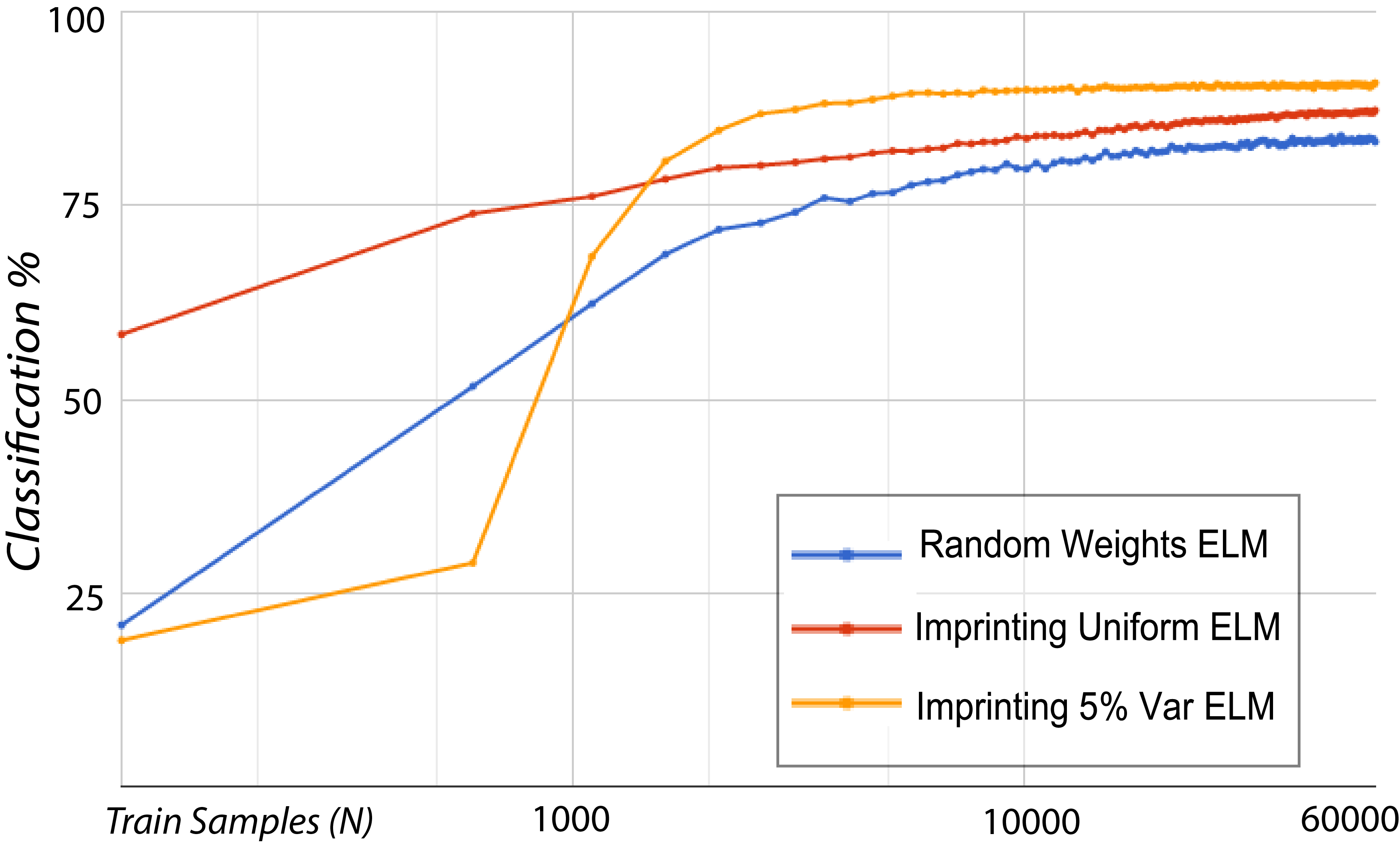}
\caption{Classification rate as a function of number of training samples $N$ used to compute \( W_{out} \) in the third stage, given $M=L=784$ (the purple line/slice in Fig.~\ref{HiddenLayer}), depicted on a log scale. Three two-layer systems cases are depicted: where \( W_{in} \) is set randomly with all ECM at low values (Random Weights ELM), and two imprinted systems where ECM cells are uniform and variable (5\% dispersion), respectively.  In imprinting cases, n=50 for uniform and n=30 for variable cases, constantly in both cases T=1s, $ \Delta t = 200 \mu s$. In all cases, 10\% pixels flipped (noise) in every image.}
\label{Training}
\end{figure}
Whether in batch or online mode, minimizing training samples number $N$ used to compose $A$ can save energy and time. Fig.~\ref{Training} shows classification rate as a function of training samples, at the case $M=L=784$. At very low sample size, the rate of improvement is high;  a steady state is reached around $N=5,000$, and performance is already within 1-2\% of maximum around $N=10,000$. 
By $N=2,500$, the variable imprinted system already outperforms the maximal result obtained for standard ELM (83.5\%); uniform imprinting surpasses by $N=8,500$. With the full training set, uniform and variable ECM projections ultimately reach new classification heights (87\% and 90\%, respectively). 
\section{Discussion}

The simple system introduced in section~\ref{sec:simple} achieves promising classification on a simple task, and does so with minimal computing accessory as patterns are remembered naturally as  a function of time and device properties. Explicit weight changes are not needed, which eliminates an impediment towards larger crossbars that require large circuit overhead for this purpose.  As memristor-CMOS systems have already been demonstrated to learn images of equivalent complexity \cite{prezioso2015training}, a physical implementation of our system is possible and could demonstrate further trade-offs. 
However, the readout involves a relatively complex procedure, and the system is sensitive to device variability.
While it is a proof of concept for harnessing transition from STP to LTP in nanodevices, it has limited applicability to real nano-electronic system design.

Conversely, the imprinted ELM architecture introduced in section~\ref{sec:ELM} is a promising lead for future nano-architectures. Imprinting a first layer with training examples definitively improves performance on the primary task. By achieving a far better classification at far smaller  hidden layer size $M$ than previously reported, the approach could dramatically reduce the total size, number of nanodevices, and CMOS neurons required to implement future ELM-inspired systems. Moreover, the fact that variable synapse ELM systems out-performed uniform synapse  ELM systems is  promising, as nanodevice variability is usually a serious concern. As nanodevices are naturally imperfect and structural synaptic diversity has been shown to enhance information coding in biological synapses \cite{bartol2015nanoconnectomic}, this implies that naturally variable filamentary nanodevices, such as our ECM cells, are excellent building blocks for future neuromorphic systems.

While gradient-based learning systems built with memristive nanodevices report $<1\%$ error on MNIST \cite{zamanidoost2015low}, to reach these heights suggested weight updates for every device must be computed externally to the system and programming pulses applied on a device-by-device basis to both layers over many epochs.  With one-shot training/testing and programming pulses only being applied to set weights on the smaller second layer (assuming $L\!\times\!M \gg J\!\times\!M$), our system might be an order of magnitude more energy efficient and  require less overhead too. Additionally, our system offers flexibility unavailable to gradient-based systems; since ridge regression solutions are iterative, low sample ($N$) solutions to \( W_{out} \) represent a trade-off between accuracy and speed/energy saving that might be intentionally exploited by approximate computing systems. Crossbars systems that use volatile memristive devices for classification were first explored in \cite{burger2014volatile}, yet percentages of $80\%+$ on the primary task were only possible when the currents of several individual crossbars were combined and when two following layers (a multi-layer perceptron), provided the solution. Our proposed system reduces area, complexity, and performance in comparison to these past schemes.
  

However, our system might fairly be considered slow due to a 'speed limit' set by device relaxation. If imprinting proceeds faster than $ \Delta t = 100 \mu S$, it oversaturates. Assuming $ \Delta t = 200 {\micro\second}$, $n=40$, $J=10$, and $T = 1s$ (MNIST), then $T_\text{tot} = ~1.1 s$ is required to imprint \( W_{in}\). Although training and testing are an order of magnitude faster, even if they were near instantaneous the system is still slower than competing nano-electronic systems. Yet, the device timing parameters used herein were academic. Device engineering, in particular device scaling, could tune $\tau_\text{fac}$ to allow for faster pattern presentation, thereby narrowing the gap between transient neuromorphic computing systems and non-volatile ones.


\section{Conclusion}

Two  novel nanoelectronic learning systems were conceived and simulated on classification tasks. 
Both systems exploit the unique properties of an ECM filamentary nanodevice with tunable STP to LTP transition to memorize and retain average images from the training sets of the classification tasks while staying immune to low levels of noise. While the simple system does well with a simple task, it can not classify the MNIST database well; moreover, variation is unfriendly to this system, and readout is complicated. For this reason, the dual-crossbar system inspired by ELM was developed to harness variability. While imprinting of the first layer provides a definitive boost over standard (random weight) ELM, the combination of synaptic and hidden layer variability help the proposed system reach $>90\%$ on the MNIST task. However, both approaches come with a fundamental timing constraint: the relaxation speed of the nanodevice implicit in the imprinting stage. While further optimization on both the device and architecture levels will be needed to reach state-of-the-art classification, benefits in terms of energy efficiency, area reduction,  partial noise immunity, and anti-fragility to synaptic variability are already apparent. 
These first results open the way for new explorations of neuromorphic architectures, which harness the intrinsic timing characteristic of nanodevices.

\section*{Acknowledgment}
This work was supported by the Nanodesign Paris-Saclay Lidex. The authors would like to thank L. Calvet, D.~Vodenicarevic, A.~Mizrahi, N.~Locatelli and J.~S.~Friedman for fruitful discussions.

\ifCLASSOPTIONcaptionsoff
  \newpage
\fi

\bibliographystyle{IEEEtran}
\bibliography{IEEEabrv,references}

\end{document}